\documentclass{article}

\usepackage{microtype}
\usepackage{graphicx}
\usepackage{adjustbox}
\usepackage{multirow}
\usepackage{subfigure}
\usepackage{booktabs} 

\usepackage{hyperref}

\usepackage[accepted]{icml2019}

\icmltitlerunning{Detecting Adversarial Examples and Other Misclassifications in Neural Networks by Introspection}

\begin{document}

\twocolumn[
\icmltitle{Detecting Adversarial Examples and Other Misclassifications in Neural Networks by Introspection}



\icmlsetsymbol{equal}{*}

 \begin{icmlauthorlist}
 \icmlauthor{Jonathan Aigrain}{AXA}
 \icmlauthor{Marcin Detyniecki}{AXA}
 \end{icmlauthorlist}

\icmlaffiliation{AXA}{AXA Services, Paris, France}

\icmlcorrespondingauthor{Jonathan Aigrain}{jonathan.aigrain@axa.com}

\icmlkeywords{Machine Learning, ICML, Adversarial, robustness, uncertainty, computer vision}

\vskip 0.3in
]



\printAffiliationsAndNotice{}  

\begin{abstract}
Despite having excellent performances for a wide variety of tasks, modern neural networks are unable to provide a reliable confidence value allowing to detect misclassifications. 
This limitation is at the heart of what is known as an adversarial example, where the network provides a wrong prediction associated with a strong confidence to a slightly modified image.
Moreover, this overconfidence issue has also been observed for regular errors and out-of-distribution data. 
We tackle this problem by what we call introspection, i.e. using the information provided by the logits of an already pretrained neural network. 
We show that by training a simple 3-layers neural network on top of the logit activations, we are able to detect misclassifications at a competitive level.
\end{abstract}

\section{Introduction}
\label{introduction}
Thanks to their excellent performances, Neural Networks (NN) are now used to tackle important problems such as medical diagnosis \cite{shen2017deep} or pedestrian detection for autonomous cars \cite{tian2015deep}. However, regarding classification, one of the issues that prevent widespread adoption of such solutions is their overconfidence in their predictions \cite{amodei2016concrete}. It has been observed that they can provide high confidence values for errors \cite{guo2017calibration}, out-of-distribution examples such as tailored noise \cite{nguyen2015deep} and adversarial examples \cite{szegedy2014}.
\par As explained in \cite{hendrycks2017baseline}, the main reason for these high confidence values is the softmax layer that is usually used as the last layer of a classification NN. As the softmax is a smooth approximation of the indicator function, it is designed to output high maximum values, even for small differences in the logits, i.e the final activation values of a classification NN before the softmax function is applied. We argue that one additional problem is that by normalizing the logits in order to obtain a probability distribution, we lose the information about their absolute values.  In this work, we focus on studying whether this information can be used to compute a confidence value that would allow to detect errors, out-of-distribution data and adversarial examples.
\par Recently, several works focused on detecting misclassifications. In \cite{hendrycks2017baseline}, the authors proposed a baseline method that uses a threshold on softmax probabilities. They also provided some guidelines about how out-of-distribution data detection experiments should be conducted and evaluated. In \cite{jiang2018trust}, the Trust Score, a specific 1-NN ratio applied on a filtered version of the training set, is presented in order to recognize misclassified examples. In \cite{liang2017enhancing}, the authors introduced \textit{ODIN}, a method that separates in and out-of-distribution examples by preprocessing the input using adversarial perturbation and then thresholding the softmax scores computed after temperature scaling.
\par In this paper, we present Introspection-Net, a simple 3 layers regression NN that takes the logits as input and aims at predicting the confidence value, i.e. whether the classification is correct (output value of 1) or not (output value of 0). We show that, by using adversarial training and data augmentation, we are able to detect misclassifications at a competitive level, as we outperform the Trust Score approach \cite{jiang2018trust} and the Softmax Baseline presented in \cite{hendrycks2017baseline}. The main contribution of this paper is to show that logits of already pretrained network provide relevant information to detect adversarial examples and other types of misclassfications.



\section{Setup and Analysis}
\label{logits}
In this Section, we show through experiments on the MNIST dataset \cite{lecunmnist} that there are significant differences in logit activations between correctly classified examples and several kinds of misclassifications.

\subsection{Baseline NN Architecture and Training}
\begin{table*}[tb]
\caption{Architecture of the NN trained on MNIST. The final accuracy on the test set is 99.65\%.}
\label{architecture-table}
\vskip 0.15in
\begin{center}
\begin{small}
\begin{sc}
\begin{tabular}{lcccccc}
\toprule
Layer type & patch size & stride & depth & padding & activation & output size\\
\midrule
Convolution & 3x3 & 1 & 32 & none & relu & 26x26x32 \\
Convolution & 3x3 & 1 & 32 & none & relu &24x24x32 \\
Max pooling & 2x2 & 2 &  &  &  & 12x12x32 \\
Dropout (20\%) &  &   &  &  &  & 12x12x32 \\
Convolution & 3x3 & 1 & 64 & same & relu & 12x12x64 \\
Convolution & 3x3 & 1 & 64 & same & relu & 12x12x64 \\
Max pooling & 2x2 & 2 &  &  &  & 6x6x64 \\
Dropout (25\%) &  &   &  &  &  & 6x6x64 \\
Convolution & 3x3 & 1 & 128 & same & relu & 6x6x128 \\
Dropout (25\%) &  &   &  &  &  & 6x6x128 \\
Flatten &  &   &  &  &  & 4608 \\
Dense &  &   &  &  & relu & 128 \\
Batch normalization &  &   &  &  &  & 128 \\
Dropout (25\%) &  &   &  &  &  & 128 \\
Dense &  &   &  &  & softmax & 10 \\
\bottomrule
\end{tabular}

\end{sc}
\end{small}
\end{center}
\vskip -0.1in
\end{table*}

For these experiments, we train a simple custom NN using the Keras framework \cite{chollet2015keras}. The architecture is described in Table \ref{architecture-table}. The training was done for 30 epochs. We used the RMSprop optimizer, a batch size of 64 and data augmentation (i.e. slight rotations, zooms and shifts). After the training, this NN achieves 99.65\% accuracy on the test set.

\subsection{Generating misclassified examples}
\label{datasets}
In order to study how logits are distributed for different kinds of misclassifications (i.e. errors, out-of-distribution and adversarial examples) compared to original MNIST images, we use or generate the following datasets:
\begin{itemize}
    \item For errors, as the performances of state of the art NNs are close to perfection on MNIST, we use the idea presented in \cite{devries2018learning}. We generate misclassified images by adding 7x7 black patches to test set images. We only keep the ones that are not classified correctly by the NN.
    \item For out-of-distribution examples, we study 4 different kinds of data. We use random normal and uniform noise images, as well as images from CIFAR-10 \cite{krizhevsky2014cifar} and Fashion MNIST \cite{xiao2017fashion} datasets.
    \item For adversarial examples, we use the CleverHans framework \cite{papernot2018cleverhans}. We generate 3 adversarial datasets using different methods: FGSM \cite{goodfellow2015}, BIM \cite{chen2019boundary} and DeepFool \cite{moosavi2016deepfool}. We only keep the images that are misclassified by the NN.
\end{itemize}

\subsection{Analysis of the difference in logit distributions}
\begin{figure}[htb]
\begin{center}
\centerline{\includegraphics[width=3.6in, height=2.5in]{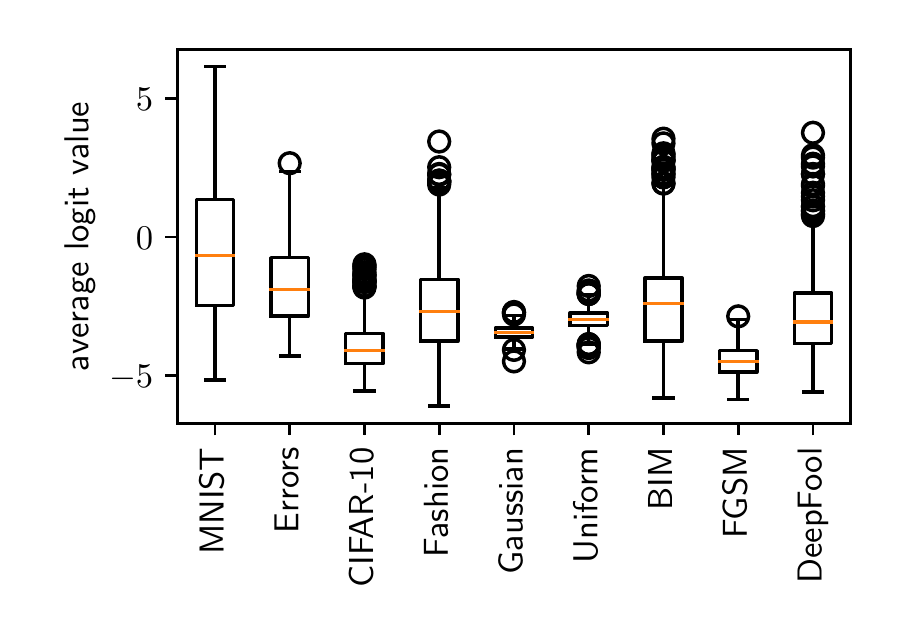}}
\caption{Distribution of the average logit values for 1000 examples of each dataset. It appears clearly that the average logit values are higher for correctly classified MNIST images compared to misclassifications.}
\label{logits-distribution}
\end{center}
\vskip -0.2in
\end{figure}

\begin{figure*}[t]
\begin{center}
\centerline{\includegraphics[width=8in, height=2.6in]{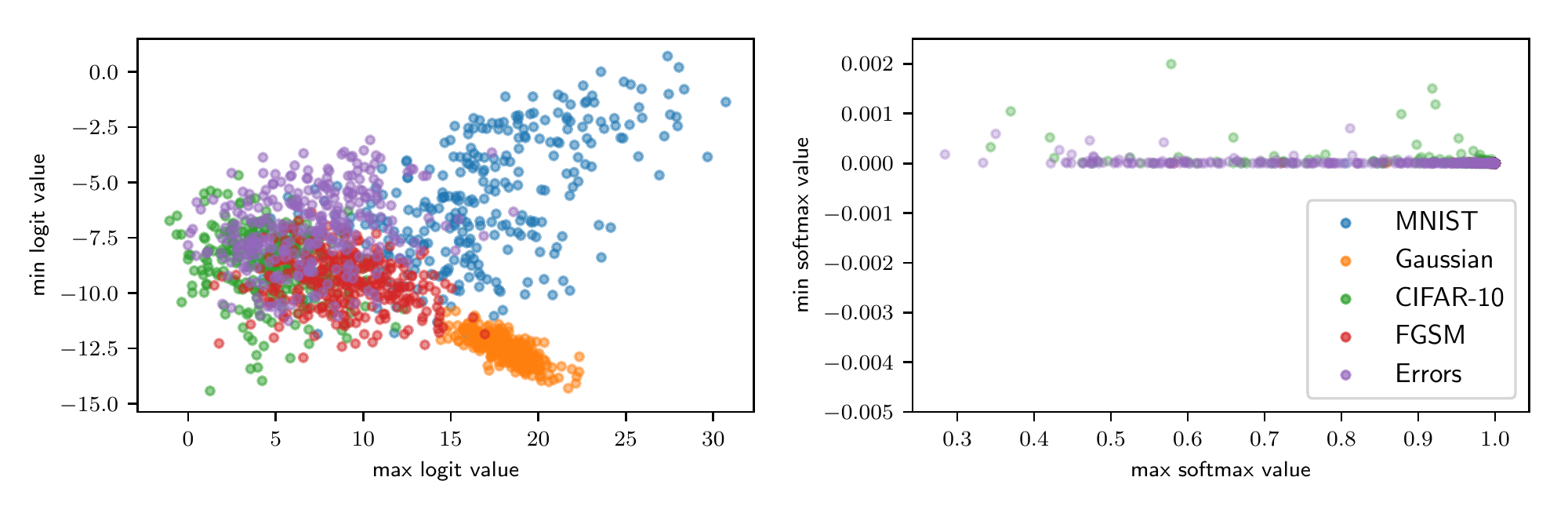}}
\caption{Scatterplot of the logit (left) and softmax (right) minimum and maximum values. For visibility reasons, we only display the values for 300 examples for 5 datasets. It appears that only using these 2 statistics for logits already helps to separate misclassifications from correct ones. However, the discriminating information is lost once the softmax function is applied. Best viewed in color.}
\label{logits-minmax}
\end{center}
\vskip -0.2in
\end{figure*}

Using the NN we described previously, we study whether logits are distributed differently among the different datasets we just presented. Figure \ref{logits-distribution} displays the distribution of the average logit values for each of them. We can clearly see that logits values are higher for correctly classified MNIST images than for any other dataset. These differences are statistically significant according to the Student's t-test ($p < 0.0001$ between MNIST and any other dataset). Intuitively, it is sound that out-of-distribution examples are associated with lower logit values. In fact, it is likely that previous convolutional layers did not detect the pixel/feature patterns they were trained on, resulting in weakly activated feature maps, which in turn leads to lower logit values. It is however interesting to note that the adversarial datasets are also associated with lower logit values. One explanation might be that in order to maximize the softmax value for a target class, it is easier to decrease logit values for the other classes than to increase the logit value of the adversarial attack target. This intuition is supported by the fact that the distribution of the logit max values is significantly higher for MNIST ($16.5\pm 5.0$) than for BIM ($10.8\pm 5.2$), FGSM ($8.8\pm 2.6$) and DeepFool ($3.4\pm 2.3$). The same observation can be made about minimum values, showing that adversarial example generation techniques tend to create images that are associated with overall lower logit values.

\subsection{Discrimination power of logit vs. softmax values}
Figure \ref{logits-minmax} presents a scatterplot of the logit and softmax minimum and maximum values for the MNIST, Gaussian, CIFAR-10, FGSM and Errors datasets. Qualitatively, it appears that these 2 simple statistics for logits are enough to separate fairly well MNIST examples from the remaining ones.  We can see that correctly classified images tend to have higher minimum and maximum logit values, which correlates well with the distributions showed in Figure \ref{logits-distribution}. However, we can see that the discriminating information is lost when we apply the softmax function. These observations confirm that logits, unlike softmax values, provide relevant information for misclassification detection. 

\section{Confidence Prediction}
\label{prediction}
\subsection{Proposed solution: Introspection-Net}


\begin{table*}[tb]
\caption{Confidence prediction results on MNIST and CIFAR-10. The best result for each metric/experience is in bold.}
\label{table-results}
\vskip 0.15in
\begin{center}
\begin{small}
\begin{sc}
\begin{tabular}{clcccc}
\hline
\textbf{Experiment}  & \textbf{OOD dataset} & \textbf{FPR (95\% TPR)}  & \textbf{AUROC} & \textbf{AUPR In} & \textbf{AUPR Out} \\ \hline
   &  & \multicolumn{4}{c}{Softmax Baseline / Trust Score / Proposed solution} \\ \cline{3-6} 
\multirow{9}{*}{\begin{tabular}[c]{@{}c@{}}Custom network\\ MNIST\end{tabular}}             & ERRORS  &\textbf{35.4}/39.5/35.8 & 86.3/86.5/\textbf{87.8} & \textbf{84.0}/82.6/82.9 & 89.6/88.9/\textbf{90.5}\\
                                & BIM    &55.1/44.7/\textbf{13.8} & 76.0/89.9/\textbf{95.8} & 75.2/90.8/\textbf{93.8} & 80.3/88.8/\textbf{96.8}\\
                                & FGSM   &68.1/69.5/\textbf{8.7} & 85.6/88.6/\textbf{97.9} & 89.6/92.5/\textbf{98.7} & 79.0/81.3/\textbf{96.0}\\
                                & DEEPFOOL   &\textbf{0.0}/0.3/2.9 & 98.1/97.9/\textbf{98.2} & \textbf{98.9}/98.8/98.9 & 96.2/95.8/\textbf{96.9}\\
                                & GAUSSIAN  &100.0/100.0/\textbf{0.0} & 24.1/82.2/\textbf{100.0} & 51.8/88.9/\textbf{100.0} & 30.0/68.0/\textbf{100.0}\\
                                & UNIFORM   &100.0/37.6/\textbf{0.0} & 21.2/95.1/\textbf{100.0} & 51.1/97.0/\textbf{100.0} & 25.0/87.8/\textbf{100.0}\\
                                & CIFAR-10   &8.8/12.6/\textbf{0.0} & 98.4/97.9/\textbf{99.9} & 98.6/98.3/\textbf{99.9} & 98.1/97.5/\textbf{99.9}\\
                                & FASHION    &23.5/21.8/\textbf{0.3} & 93.7/96.0/\textbf{99.8} & 92.8/96.1/\textbf{99.8} & 94.6/96.0/\textbf{99.8}\\ \hline
\multirow{9}{*}{\begin{tabular}[c]{@{}c@{}}WR-28-8\\ CIFAR-10\end{tabular}}       & ERRORS  &78.2/82.9/\textbf{75.4} & 67.6/65.8/\textbf{69.8} & \textbf{67.7}/62.1/65.8 & 68.5/66.6/\textbf{71.4}\\
                                & BIM    &99.9/99.7/\textbf{7.3} & 14.4/56.1/\textbf{97.9} & 42.7/62.6/\textbf{96.7} & 28.6/49.5/\textbf{98.4}\\
                                & FGSM   &71.5/71.2/\textbf{55.7} & 89.5/90.7/\textbf{94.4} & 93.7/94.8/\textbf{97.0} & 80.6/80.4/\textbf{87.4}\\
                                & DEEPFOOL   &\textbf{50.8}/55.8/71.6 & \textbf{95.0}/91.0/89.0 & \textbf{97.2}/93.9/93.1 & \textbf{89.3}/85.4/81.5\\
                                & GAUSSIAN  &24.6/9.2/\textbf{0.0} & 96.1/98.4/\textbf{99.9} & 97.0/98.7/\textbf{100.0} & 95.2/97.7/\textbf{99.8}\\
                                & UNIFORM   &89.4/33.2/\textbf{0.0} & 85.6/95.4/\textbf{99.9} & 90.4/96.7/\textbf{100.0} & 74.9/91.9/\textbf{99.8}\\
                                & MNIST    &47.6/65.4/\textbf{0.9} & 92.6/89.0/\textbf{99.7} & 94.5/91.4/\textbf{99.7} & 90.1/85.3/\textbf{99.7}\\
                                & FASHION    &46.5/50.1/\textbf{3.3} & 92.4/90.7/\textbf{99.1} & 94.0/91.9/\textbf{99.0} & 90.5/88.9/\textbf{99.1}\\ \hline
\end{tabular}

\end{sc}
\end{small}
\end{center}
\vskip -0.1in
\end{table*}

Based on the insights revealed in the previous section, we train a simple 3 layers regression NN which we call Introspection-Net since it takes an intermediate layer, the logits, as input. It aims at predicting the confidence value associated to a given prediction, i.e. whether it is a correct (value of 1) or incorrect (value of 0). Introspection-Net is composed of 3 dense layers with 128 neurons and RELU activations. The first 2 layers are followed by dropout layers with a dropout rate of 20\%, and the second dropout layer is also followed by a Batch Normalization layer. We train the network for 60 epochs using RMSProp to optimize the mean squared error loss. 


\subsection{Experimental setup}
\textbf{Experiments:} To evaluate our proposition we run the following two experiments. In a first one, we predict the confidence values associated to the predictions made by the NN we described in the previous Section. 
In a second one, we predict the confidence values associated to the predictions of a Wide Residual Network \cite{zagoruyko2016}, with depth 28 and width 8, trained on CIFAR-10. We use the implementation provided by Keras contrib \cite{chollet2015keras}. We run this second experiment to ensure that our approach can generalize to several datasets and NN architectures.
For both experiments, we compare 3 methods: the Softmax Baseline introduced in \cite{hendrycks2017baseline}, the Trust Score \cite{jiang2018trust} and our method. 
\par\textbf{Evaluation metrics:} We use the evaluation framework introduced in \cite{hendrycks2017baseline}: the Area Under the Receiver Operating Characteristic curve (AUROC), the Areas Under the Precision-Recall curve (AUPR) for both the correct (AUPR In) and incorrect (AUPR Out) classes. In addition, we also compute the False Positive Rate at 95\% True Positive Rate as in \cite{liang2017enhancing}.
\par\textbf{Data:} For both experiments, the training set is composed of 5000 examples belonging to the in-distribution dataset (i.e. MNIST for experiment 1, and CIFAR-10 for experiment 2), and of 700 examples of all the remaining datasets presented in Section \ref{datasets} (adversarial examples and errors are generated using CIFAR-10 images for experiment 2). These numbers were chosen in order to have a balanced dataset. From this data, we use 10\% as a validation set. The testing set is composed of 2000 images of each dataset.

\subsection{Results}
The experimental results are shown in Table \ref{table-results}. We can see that our method provides overall significantly better confidence values than the other methods, regardless of the in-distribution data and of the chosen NN architecture. It is however important to keep in mind that the Trust Score has the advantage to be model-agnostic, unlike our solution.
\par Regarding the error detection task, all 3 methods obtain fairly similar results. It is the most challenging task for our method, as it achieves its worst performances for both experiments. 
\par Our method performs extremely well on the out-of-distribution data detection task, as we obtain close to perfect scores for all datasets (i.e. Gaussian, Uniform, CIFAR-10/MNIST and Fashion) in both experiments. These results are a significant improvement over the two other methods. For instance, in the first experiment we achieve a 0.0\% FPR on the Gaussian dataset while both the Trust Score and the Softmax Baseline obtain a 100.0\% FPR.
\par Our method also achieves overall better results for adversarial example detection. It is especially striking for the BIM dataset, as we achieve a 97.6 AUROC in the second experiment compared to 14.4 for the Softmax Baseline and  56.1 for the Trust Score. However, the Baseline Softmax provide the best results for detecting DeepFool adversarial examples on CIFAR-10. This is due to the fact that softmax values are highly discriminative for these examples, as the average maximum softmax probability is only $54.0\%$ for DeepFool and $94.8\%$ for correctly classified MNIST images. 

\section{Conclusion and Perspectives}
We have shown through a series of experiments that logits, unlike softmax probabilities, of already pretrained neural networks provide relevant information to detect 3 types of misclassifications: errors, out-of-distribution data and adversarial examples. We have proposed Introspection-Net, a neural network trained on logit activations to predict whether a prediction is correct or not. This solution outperformed by a large margin the Softmax Baseline and the Trust Score on confidence prediction, without requiring to retrain the original NN. 
\par Our findings highlight the interest of using Introspection, i.e. using NN learned internal representations, to detect misclassifications. These results are especially interesting in the case of adversarial examples detection, since they show that, although the softmax values are "fooled" by the adversarial noise, it is not the case for internal representations such as logits, even without any additional training procedure.
\par On the other hand, Inspection-Net does require adversarial training to learn the logit distributions for different types of misclassifications, which is one of its main drawbacks. Consequently, future work involves studying whether one can detect these misclassifications using only the logits of in-distribution examples. We are also interested in exploring whether this method can be used for other applications such as natural language processing.




\bibliography{paper}
\bibliographystyle{icml2019}


\end{document}